# Efficient task and path planning for maintenance automation using a robot system

Christian Friedrich, Akos Csiszar, Armin Lechler, and Alexander Verl

*Abstract*—The research and development of intelligent automation solutions is a ground-breaking point for the factory of the future. A promising and challenging mission is the use of autonomous robot systems to automate tasks in the field of maintenance. For this purpose, the robot system must be able to plan autonomously the different manipulation tasks and the corresponding paths. Basic requirements are the development of algorithms with a low computational complexity and the possibility to deal with environmental uncertainties. In this work, an approach is presented, which is especially suited to solve the problem of maintenance automation. For this purpose, offline data from CAD is combined with online data from an RGBD vision system via a probabilistic filter, to compensate uncertainties from offline data. For planning the different tasks, a method is explained, which use a symbolic description, founded on a novel sampling-based method to compute the disassembly space. For path planning we use global state-of-the-art algorithms with a method that allows the adaption of the exploration stepsize in order to reduce the planning time. Every method is experimentally validated and discussed.

*Note to Practitioners*—This paper was motivated by the current deficit to automate maintenance tasks in manufacturing systems. Today there exist many automatic methods for fault detection, but there is no possibility to restore the desired state. A typical application is the replacement of a defective component or service tasks like refilling of cooling lubricant in machine tools. The focus of this work is to provide automatic planning concepts for robot systems which deals with this problem. For this, a task planning algorithm is proposed which uses CAD and vision data. The advantage of this method is that it works on a contact analysis, based on a discrete approximation of the disassembly space, allowing a fast determination of possible robot manipulations. Also a new pre-processing concept for path planning is developed which enables the adaptation of the stepsize of the path planner, based on the local occupancy of the environment. Through this, it is possible to drastically reduce the required planning effort.

## I. INTRODUCTION

IN today's industrial context, the availability of manufacturing systems plays an increasingly important role in maximizing the factory's economics. Studies show that the maintenance costs amount up to 35% [1] of the total life cycle costs for machines tools. For supporting the maintenance staff and improving the factories economics, the development of automation solutions is a key factor.

Manuscript received February 6, 2017. Research supported by the German Research Foundation (DFG). Christian Friedrich, Akos Csiszar, Armin Lechler, and Alexander Verl are with the Institute of Control Engineering of Tool Machines and Manufacturing Units (ISW), University

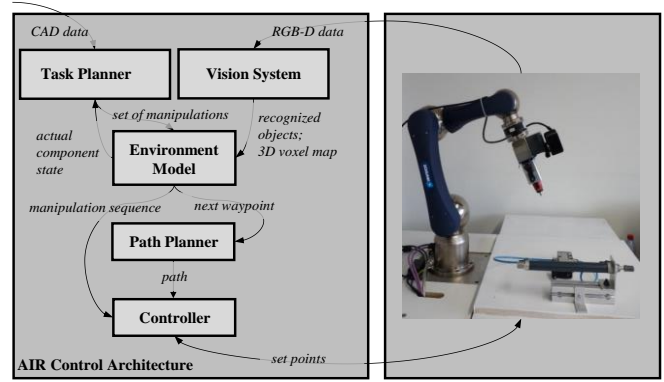

Fig. 1. General architecture of the Autonomous Maintenance Robot.

While the use of automation solutions has been extensively discussed for fault diagnosis [2], [3], there is a technology gap in systems for restoring the intended plant or machine states. For the automation of repair or service tasks, the use of robotic systems is common, especially in areas inaccessible for humans. However, these are developed mostly for specific areas, such as energy transmission networks [4], [5], sewer systems [6], [7] or thermal boilers [8]. Also for complex areas like bridges [9], [10] material flow systems [11] or process plants [12], [13], the use of robot systems was investigated. However, the state of the art discusses no generalized approach to plan and execute different manipulation tasks for maintenance automation. Typical applications are the exchange of broken components in a manufacturing system or different service tasks like cleaning. Figure 1 shows the used system for this work.

The goal of this paper is to present a fundamental approach for task and path planning, in order to enable the automatic execution of maintenance tasks by a robotic system. Furthermore, we propose a method for combining the symbolic oriented task with the path planning component.

## II. STATE OF THE ART

In this section, we will discuss the most relevant work in related domains like (dis-)assembly automation. For a general overview see [14], [15].

The field of assembly automation has a long tradition in robotics. To model mechanical assemblies, usually methods from graph theory are used, where vertices are the components and edges the connection between them. Commonly used representations are the Liaison-Graph [16] or more extended methods like the relation graph [17], which is also used in this approach. Through so-called symbolic spatial relations [18], which represent the contact between components, the symbolic degree of freedom can be principally determined.

To model disassembly or assembly sequences there are different approaches, which use directed graphs, AND/OR-graphs [19], diamond diagrams [20], or petri nets [21]. The work [22] discusses the theoretical aspects in relation to completeness and correctness of the different representations. In general, the calculation of an optimal disassembly sequence is NP-complete [23], which places high demands on the design of appropriate heuristics [24]. For computing a possible sequence, different methods are known in literature, like the Directional and Non-Directional Blocking Graph [25], the Disassembly Precedence Matrix [21], [23], or the cut-set method [26]. The cut-set method decomposes the connection graph into all possible subgraphs, which is in general an assembly-by-disassembly strategy. This approach is still the basis for many advanced and new methods.

One of the most sophisticated assembly planning systems is $^{\text{High}}$Lap (High Level Assembly Planning) [17], [27], [28]. The system combines planning and execution of the assembly tasks by means of a robot system and uses for this purpose CAD data and user assistance. The works [29], [24], [30], extend the system by using more efficient algorithms to solve the geometric decomposition. These methods are extremely suitable for assembly automation, because the computation of a nearly optimal assembly sequence is provided. A disadvantage of this method for maintenance automation is that the planning is based purely on geometric CAD data, which makes it time-consuming and unable to compensate deviations from the real environment. Another problem is that the task is directly decomposed into skill primitives [31], so that only pre-specified tasks can be solved. The use of more symbolic definitions of robot skills above the skill primitive layer could solve this problem.

Solutions in disassembly automation, which are used for recycling tasks, usually combine real environment information from vision with offline available product data, compare, [32], [33], and [34]. A drawback of all these solutions is that they do not provide an embedded scheme for sequence and path planning. In addition, they discuss no general method of combining the offline with the online data from vision and also provide no solution in regards to handling classification errors caused by vision algorithms.

After task planning, the corresponding path must be determined. In principle, the topic of path planning is regularly discussed in robotics and there exist several monographies [35], [36], which cover the topic. The methods can be divided into local, global and hybrid path planning approaches. Global approaches, which are especially well suited for our approach, use search- or sampling-based methods to find a collision-free path. Search-based planning methods like A$^*$ [37] and anytime algorithms like ARA$^*$ [38] are complete and optimal if a solution exists [39]. Sampling-based planning such as Probabilistic Roadmaps (PRM) or Rapidly-Exploring Random Trees (RRT) and more extended variants [40], [41] are often better suited for high-dimensional problems and are probabilistically complete [35]. A global approach is also considered in this work, because during the execution of the different tasks a static environment is assumed. In addition, it makes it easier to combine task and path planning through the used environment model. However, a big problem of global path planning is that the computational time scales differ depending on the occupancy of the environment. In order to deal with this problem, a method is required to adapt the exploration stepsize of the path planning algorithms automatically for reducing the planning time and increasing the success rate for finding a possible path. To provide a general solution in maintenance automation, the developed methods should satisfy the following requirements:

- Task and path planning algorithms, with a low computational complexity to find the required manipulations for solving a maintenance task.
- Fusion of passive (CAD data) and active information (vision data) in order to incorporate environmental uncertainties.
- Coupling of the symbolic task planning with the numerical path planning into a general control concept.

### III. AUTOMATIC TASK AND PATH PLANNING FOR MAINTENANCE AUTOMATION

For the automation of maintenance tasks through a robot system the robot must be able to autonomously plan the different tasks and the corresponding paths. Therefore, we suggest novel solutions for task and path planning, which are adapted to the specific requirements mentioned in section II. Further, we provide a general solution for compensating environmental uncertainties via probabilistic filtering and a method to interface the symbolic task with the path planning in a general control scheme.

#### A. Task planning

Previous work of the task planning approach was also discussed in [42], and is here only explained for a better understanding of the complete system. For the design of the task planner, we use an extended version of the relational assembly model $\mathcal{RM}$ [17]. The relational assembly model contains information about the symbolic spatial relations $SSR \in \{concentric, congruent, screwed\}$ between two components $C$, which are defined on a feature geometry $\mathcal{FG}_{SSR} \in \{plane, line, point, circle\}$. For all $SSR$ also a feature vector $\underline{f}_{SSR} = [\underline{o}_{SSR} \; \underline{e}_{SSR}]^T$ is available with an origin $\underline{o}_{SSR} = [o_x \; o_y \; o_z]^T$ and a normal vector $\underline{e}_{SSR} = [e_x \; e_y \; e_z]^T$ at the origin. The relations are stored in a graph $\mathcal{G}_{SSR}(C, SSR)$. Therefore, the required data from the CAD assembly model is imported into the planner. Furthermore, the relation graph is generated using a new sampling-based approach to compute the relative degrees of freedom between all components. The resulting graph can be further used to plan the required actions based on the concept of manipulation primitives. This results in a set of topological executable manipulation primitives, which are further sequenced with respect to the absolute degree of freedom of a component.

*1) Hybrid manipulation planning*

The hybrid manipulation planner (HMP) is divided into a pre-processor to compute the relation graph $\mathcal{G}_{sdof}$ and in a processor to find the required set of manipulations $\{\mathcal{MP}\}$ to solve a task. The relation graph defines the components $C$ as vertices and the edges between them as the symbolic degree of freedom $sdof \in \{fix, lin, rot, fits, agpp, free\}$ [17]. To compute the relation graph $\mathcal{G}_{sdof}$ we introduce a new concept, based on a sampling-based geometric analysis. A sampling-based approach is considered, because it allows a fast computation of the possible disassembly space $\mathbb{W}_D$.

Therefore, a unit sphere $\mathcal{S} = \{\underline{x} \in \mathbb{R}^3 \mid \|\underline{x}\|_2 = 1\}$ is computed in a discrete manner using the Marsaglia method [43]. The sphere $\mathcal{S}$ represents the complete disassembly space $\mathbb{W}_D$ of all translational disassembly directions for an unconnected component. Further, for every feature geometry $\mathcal{FG}_{SSR,j}$ which is connected from a component $C_i$ to other components, the disassembly space $\mathbb{W}_{D,i} = \mathbb{W}_D(\mathcal{FG}_{SSR,j})$, $\forall j$ is computed. For example, the disassembly space of a congruent connection on a plane is described as a hemisphere. The computation is done using the parameterization of the feature geometry, which is further rotated in the direction of the feature vector $\underline{f}_{SSR}$. To compute the resulting disassembly space of $C_j$ equation (1) is considered.

$$\mathbb{W}_{D,i} = \left(\bigcap_{j=1}^{M} \mathbb{W}_D(\mathcal{FG}_{SSR,j})\right) \cap \mathbb{W}_D \quad (1)$$

The evaluation of (1) becomes straightforward. Because all geometric elements are represented as a discrete set, the values are sorted and compared to get the intersection set. So the complexity of computing $\mathbb{W}_{D,i}$ is in $\mathcal{O}(n \cdot \log(n))$. The symbolic relation $sdof$ is determined based on the resulting disassembly space of the component. Algorithm 1 and Figure 2 summarize this approach.

---

**Algorithm 1:** Sampling-based DoF Analysis (SDA)

**Input:** relational assembly model $\mathcal{G}_{SSR}(C, SSR)$.
**Output:** disassembly space $\langle\mathbb{W}_{D,C}\rangle$; relation graph $\mathcal{G}_{sdof}(C, sdof)$.

1: // Compute sampled sphere with N-points
2: $\mathcal{S} \leftarrow \text{marsagliaSphere}(n)$
3: $\mathbb{W}_{D,i} \leftarrow \mathcal{S}$;
4: **for** $i = 1:1:|C|$
5:    **for** $j = 1:1:|\mathcal{FG}_{SSR}|$
6:       $\mathbb{W}_{D,j} \leftarrow \text{getSubSpace}(\mathcal{FG}_{SSR,j}, \underline{f}_{SSR})$;
7:    **end for**
8:    $\mathbb{W}_{D,i} \leftarrow \text{intersect}(\mathbb{W}_{D,j}, \mathbb{W}_{D,i})$;
9:    $\langle\mathbb{W}_{D,C}\rangle \leftarrow \mathbb{W}_{D,i}$;
10: **end for**
11: $\mathcal{G}_{sdof} \leftarrow \text{getRelationalGraph}(\langle\mathbb{W}_{D,C}\rangle, \mathcal{G}_{SSR})$;
12: **return** $(\langle\mathbb{W}_{D,C}\rangle, \mathcal{G}_{sdof})$;

1: **function** $(\mathcal{S}) \leftarrow \text{marsagliaSphere}(n)$
2:   $i = 1$;
3:   **while** $(i \leq n)$
4:     // Get two random numbers from a uniform distribution
5:     $\omega_1 \leftarrow \mathcal{U}(-1,1)$;
6:     $\omega_2 \leftarrow \mathcal{U}(-1,1)$;
7:     **if** $(\omega_1^2 + \omega_2^2 < 1)$ **then**
8:       $x(i) \leftarrow 2\omega_1\sqrt{1 - \omega_1^2 - \omega_2^2}$;
9:       $y(i) \leftarrow 2\omega_2\sqrt{1 - \omega_1^2 - \omega_2^2}$;
10:      $z(i) \leftarrow 1 - 2(\omega_1^2 + \omega_2^2)$;
11:      $i = i + 1$;
12:     **end if**
13:   **end while**
14:   $\mathcal{S} \leftarrow [x(:)\ y(:)\ z(:)]$;
15: **end function**

---

After the computation of the relation graph $\mathcal{G}_{sdof}(C, sdof)$, the hybrid manipulation planner computes the different manipulations using a symbolic description directly in the robot task space. The common goal is to disassemble a broken component $C_{da}$, so that $\mathcal{G}_{sdof} \cap C_{da} := \emptyset$. For this purpose, we introduce the concept of manipulation primitives $\mathcal{MP}$. Here, manipulation primitives are a symbolic description, which describes the skills of the robot. The skills are defined by the following concepts:

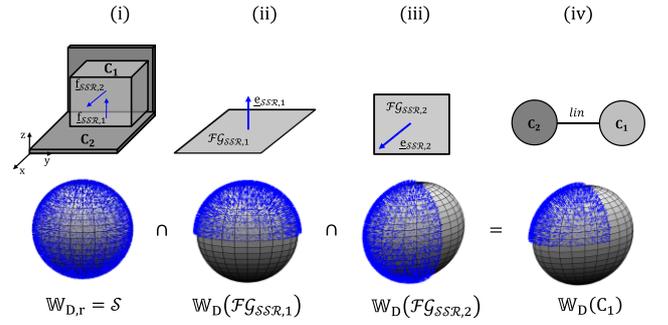

Fig. 2. Sampling-based DoF analysis: sampling the sphere to get the initial disassembly space $\mathbb{W}_D$ without any defined relation (i); sampling of the feature geometries $\mathcal{FG}_{SSR,1}$ (ii) and $\mathcal{FG}_{SSR,2}$ (iii) to get the disassembly space of the defined symbolic spatial relation; computation of the resulting disassembly space of component $C_1$ using sort and compare operations (iv).

***move***($\cdot$): *Describes a movement from the symbolic pose $p_i$ to the symbolic pose $p_{i+1}$ in the reference frame $\mathcal{F}_\mathcal{R}$.*
***twist***($\cdot$): *Describes a rotatory movement from the symbolic pose $p_i$ to the symbolic pose $p_{i+1}$ in the reference frame $\mathcal{F}_\mathcal{R}$, with a required symbolic force/torque $f_i$.*
***pull***($\cdot$): *Describes a translational movement from the symbolic pose $p_i$ to the symbolic pose $p_{i+1}$ in the reference frame $\mathcal{F}_\mathcal{R}$, with a required symbolic force/torque $f_i$.*
***put***($\cdot$): *Describes a translational movement from the symbolic pose $p_i$ to the symbolic pose $p_{i+1}$ in the reference frame $\mathcal{F}_\mathcal{R}$, with a required symbolic force/torque $f_i$.*

To compute the required manipulation primitives and to solve the task, a so-called symbolic disassembly logic is introduced. The symbolic disassembly logic describes the state transitions $\delta: \mathcal{Z} \times \Sigma \rightarrow \mathcal{Z}$, for a set of states $\mathcal{Z} := sdof_j(C_i), \forall i \in \mathcal{G}_{sdof}$, which is dependent upon the applied actions $\Sigma := \{\mathcal{MP}\} \in \{move, twist, pull, put\}$. So the state transition is given by

$$sdof_{k+1}(C_i) := \delta(sdof_k(C_i), \mathcal{MP}_k). \quad (2)$$

To apply a manipulation primitive to a component, also a tool $\tau_j$ is required. So we can rewrite (2) and define it by the rule (3) that the state $sdof_k(C_i)$ is changed to $sdof_{k+1}(C_i)$ if a manipulation primitive $\mathcal{MP}_k$ is applied to $C_i$ with a tool $\tau_j$.

$$sdof_{k+1}(C_i) \Leftarrow \mathcal{MP}_k[\tau_j, sdof_k(C_i)]. \quad (3)$$

To update the resulting $sdof_{k+1}$, the processor uses a knowledge base with basic patterns or sends a query to the pre-processor to update $\mathcal{G}_{sdof}$. A rule-based inference is used for the choice of the manipulation primitive (4) which depends on the actual state of the component and also to derive the tool (5).

$$sdof_k(C_i) \Rightarrow \mathcal{MP}_j. \quad (4)$$

$$C_i \wedge \mathcal{MP}_j \Rightarrow \tau_j. \quad (5)$$

The result is a set of manipulation primitives $\{\mathcal{MP}\}$, which are only executable if the relative disassembly space is equal to the absolute, because the planner respects only the relative degree of freedom. To sequence the set in an executable way, section B introduces a method based on visibility and accessibility of the components.

*2) Experimental results: task planning*

The HMP is evaluated using different academic and application-oriented examples. In Figure 3 the pre-processing approach is shown using Algorithm 1. After sampling the sphere for every component, the disassembly space is computed. TABLE I shows the results for different assembly groups. All experiments are done using an Intel Core i5-6200U, 2.3GHz with 8GB RAM. The metrics used are:

- pre-processing time $t_{pre}$ in [s] per component
- processing time $t_{pro}$ in [s] per component
- set of required manipulation primitives $\{\mathcal{MP}\}$

$|C|$ represents the amount of components and $|sdof|$ the number of connections per assembly. Since no collision analysis is required to compute the disassembly direction of the components, the planning can be rapidly done using only the contact information and consider only the relative degree of freedom. The task planning is further done in a purely symbolic way to generate the required manipulations.

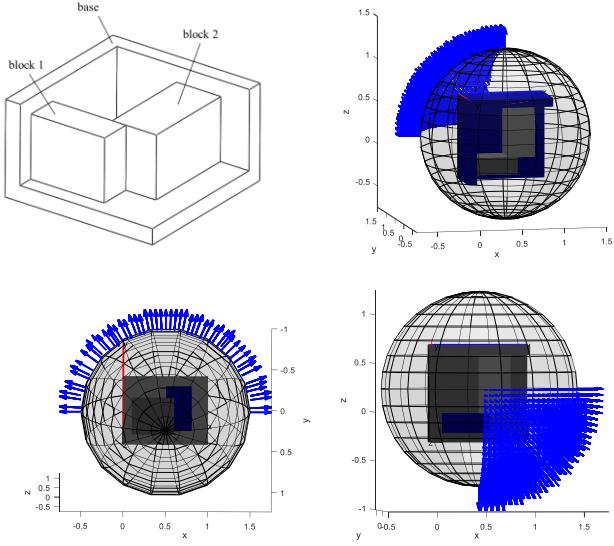

Fig. 3. Pre-processing approach: CAD input data (left; top); resulting disassembly direction in which the components can be moved in the disassembly space for the base (right; top), block 2 (left; down) and block 1 (right; down).

TABLE I. EVALUATED METRICS FOR TASK PLANNING

| Assembly Group | | |
|---|---|---|
| $|C| \in \mathbb{N}_+$ | 2 | 5 |
| $|\mathcal{SDoF}| \in \mathbb{N}_+$ | 1 | 10 |
| $C_{da}$ | block | puzzle 1 |
| $t_{pre}$ in [s] | 0.058 | 0.0581 |
| $t_{pro}$ in [s] | 0.004 | 0.0062 |
| $\{\mathcal{MP}\}$ | free(block) ⇐ pull[gr.2f, lin(block)] | free(puzzle 1) ⇐ pull[gr.2f, lin(puzzle 1)] |
| Assembly Group | | |
| $|C| \in \mathbb{N}_+$ | 4 | 9 |
| $|\mathcal{SDoF}| \in \mathbb{N}_+$ | 5 | 11 |
| $C_{da}$ | block | valve |
| $t_{pre}$ in [s] | 0.039 | 0.0376 |
| $t_{pro}$ in [s] | 0.007 | 0.0101 |
| $\{\mathcal{MP}\}$ | fits(screw_1) ⇐ twist[sd, rot(screw_1)]<br>free(screw_1) ⇐ pull[gr.3f, fits(screw_1)]<br>fits(screw_2) ⇐ twist[sd, rot(screw_2)]<br>free(screw_2) ⇐ pull[gr.3f, fits(screw_2)]<br>free(block) ⇐ pull[gr.2f, aggp(block)] | free(airhose_1) ⇐ pull[gr.ah, fits(airhose_1)]<br>free(airhose_3) ⇐ pull[gr.ah, fits(airhose_3)]<br>fits(screw_1) ⇐ twist[sd.M4, rot(screw_1)]<br>free(screw_1) ⇐ pull[gr.3f, fits(screw_1)]<br>free(connector) ⇐ pull[gr.2f, lin(connector)]<br>fits(screw_2) ⇐ twist[sd.M4, rot(screw_2)]<br>free(screw_2) ⇐ pull[gr.3f, fits(screw_2)]<br>fits(screw_3) ⇐ twist[sd.M4, rot(screw_3)]<br>free(screw_3) ⇐ pull[gr.3f, fits(screw_3)]<br>free(valve) ⇐ pull[gr.2f, aggp(valve)] |

The average planning time is 0.05s for the computation of a manipulation primitive (pre- and processing). The disadvantage of this method is, of course, that the information about the contact states is required. Another drawback at this moment is that we can only generate linear monotonous plans. If a single component has no relative degree of freedom, the planner cannot combine different components to a sub-assembly to find a possible disassembly space. Also, for the use of a dual-arm kinematics the planning algorithm must be extended to a broader planning class.

### B. Data fusion and metric assignment

To fuse the passive information from CAD with the active information from vision, we apply a probabilistic approach. This provides a way to fuse the information and to compensate for the environmental uncertainty. Based on this a possibility is shown, which allows time efficient sequencing of the given set of manipulation primitives $\{\mathcal{MP}\}$, which is evaluated using different heuristics.

*1) Probabilistic information fusion*

To update the degree of belief about the environment information, we apply a probabilistic model $P(C_i|z_t)$. So the computation of the degree of belief $bel(C_{i,t})$ that a component $C_i$ is on the position $r_{C,i}$ can be updated using a binary Bayes-Filter (6), makes the fusion straightforward.

$$\underbrace{\frac{P(C_i|z_{1:t})}{1-P(C_i|z_{1:t})}}_{bel(C_{i,t})} = \underbrace{\frac{P(C_i|z_t)}{1-P(C_i|z_t)}}_{S_{C_i}} \cdot \underbrace{\frac{P(C_i|z_{1:t-1})}{1-P(C_i|z_{1:t-1})}}_{bel(C_{i,t-1})}, \quad (6)$$

with the initial degree of believe $bel(C_{i,0}) = bel(C_i^-)$. For each component a separate binary Bayes-Filter is allocated, so

$$\begin{pmatrix} bel(C_{0,t}) \\ bel(C_{1,t}) \\ \vdots \\ bel(C_{N,t}) \end{pmatrix} = diag(S_{C_0}, S_{C_1}, \dots, S_{C_N}) \cdot \begin{pmatrix} bel(C_{0,t-1}) \\ bel(C_{1,t-1}) \\ \vdots \\ bel(C_{N,t-1}) \end{pmatrix}. \quad (7)$$

For a defined threshold $bel(C_{i,t}) \geq P_{trust,i}$, the information state is assumed to be the truth and is accepted by the environment model.

*2) Visibility and accessibility space*

To compute an ordered sequence of manipulation primitives $\langle \mathcal{MP} \rangle$ from the set $\{\mathcal{MP}\}$, we introduce the visibility $\mathcal{W}_V$ and the accessibility space $\mathcal{W}_A$. This means that all components $C_i$, which are visible at the actual disassembly state, can be manipulated, because there exists a subset $\{\mathcal{MP}_{vis}\} \subseteq \{\mathcal{MP}\}$. The visibility of $C_i$ results from the states of (7). Further, it is tested whether the components are accessible at the actual disassembly state, which results in a further subset $\{\mathcal{MP}_{acc}\} \subseteq \{\mathcal{MP}_{vis}\} \subseteq \{\mathcal{MP}\}$. The resulting subset $\{\mathcal{MP}_{acc}\}$ can be further used for task sequencing, compare Figure 4.

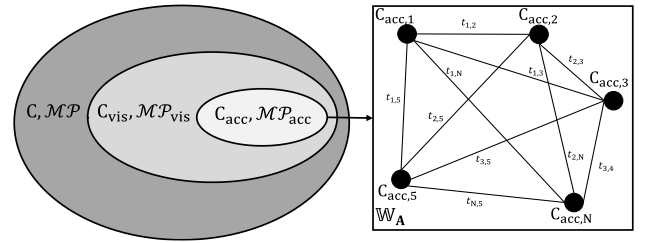

Fig. 4. Concept of visibility and accessibility space for task sequencing.

This method allows only to consider a subset of $\{\mathcal{MP}\}$ and not to find the global optimal sequence. But with this method it is possible to rapidly compute a local optimal sequence over $\mathcal{W}_A$, because only one collision analysis for the interaction of the robot with the component is required.

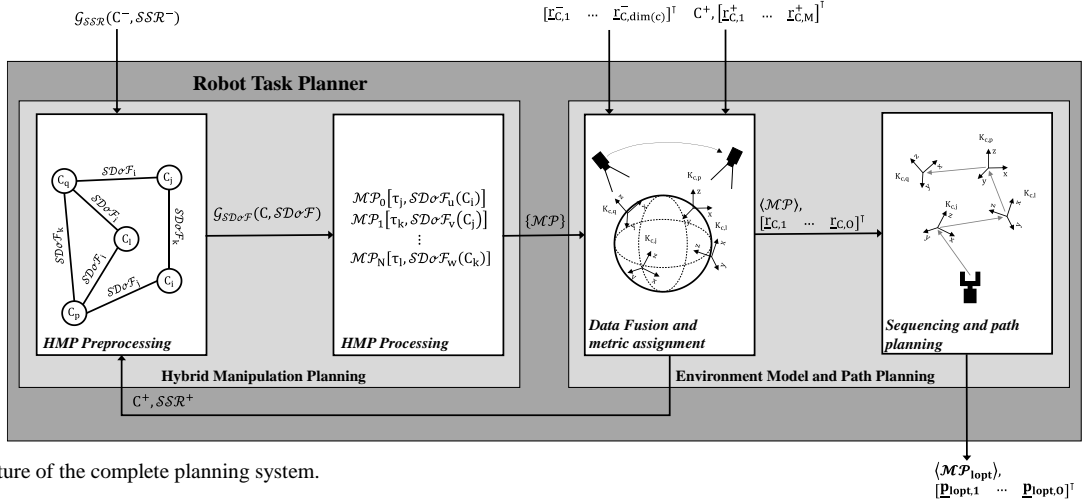

Fig. 5. Architecture of the complete planning system.

### C. Task sequencing

After the manipulation planning a list of manipulation primitives are available. All of these have to be carried out by the robot in order to complete the task. However, the order in which these are carried out is not strictly defined. In order to assure the efficiency of the maintenance process, the sequence of the manipulation primitives has to be optimized for minimum overall execution time, as presented in (8).

$$t^* = \arg\min_{t_{path}, t_\tau \in D} \langle \mathcal{MP} \rangle. \quad (8)$$

As it can be seen in Figure 4, the manipulation primitives can be represented as a graph structure and the transition times (travel time $t_{path}$ and tool change time $t_\tau$) can be represented as edges costs in the graphs. The problem with this solution is similar to a Travelling Salesman Problem (TSP) for which many solutions are well studied. The approach selected here is based on the A$^*$ algorithm, since it is already used for path planning in this project. A$^*$ is not the most typical approach for solving TSP problems, and so it might not be the most efficient way but it is effective (returns the optimal results). Furthermore, the proposed concept is not dependent on one given optimization method, this is why it was interesting to explore the differences between different types of heuristics and also highly efficient methods (like the nearest-neighbor method), which do not guarantee an optimal solution. In order to use A$^*$ for sequencing, the stop condition has been changed. Instead of reaching a goal state, the stop condition is reaching all required states. The only way the runtime of the A$^*$ algorithm can be significantly improved is through the use of heuristics. To compare the efficiency, the following three heuristics have been used.

- *Zero heuristics (Dijkstra):* Using a constant zero heuristic the A$^*$ algorithm is transformed in the Dijkstra algorithm. The predicted cost of reaching the target is constantly zero.
- *Nearest (unvisited) neighbor heuristics (A$^*$-NN):* In this second case, the predicted cost for reaching the goal is equal to the cost of transitioning to the nearest-neighbor node in the graph, which has previously not been visited in the solution candidate.
- *Minimum spanning tree heuristics (A$^*$-MST):* The third heuristics implemented for comparison of running times is based on selecting the total cost of the minimum subtree, formed in between the nodes of the graph not yet visited in the candidate solution.

Also a basic nearest-neighbor (NN) algorithm is applied.

#### 1) Experimental results: task sequencing

The task sequencing has been carried out for a sample set of four, five and six manipulation primitives. All experiments are repeated 100 times with an initialized random cost matrix. Figure 6 shows the results of the task sequencing process.

Using the A$^*$ with valid heuristics guarantees an optimal solution but has bad scaling of the planning time. On the other hand the basic nearest-neighbor algorithm scales well with the amount of manipulation primitives but guarantees no optimal solution. The improvement of the task executing time $t^*$ using A$^*$ with heuristics is in average approximately 21% against the use of the nearest-neighbor solution, compare Figure 6, right. The disadvantage of the optimal solution is of course the scaling of the planning time. Using four manipulation primitives in contrast to six results in this case using Dijkstra in a 9.3 times higher planning time $t_p$. Using A$^*$-NN results in a 46.8 times and for A$^*$-MST in a 10.2 times higher planning time, see Figure 6, left. Because the nearest-neighbor solution scales in $\mathcal{O}(N^2)$, with $N = |\langle \mathcal{MP} \rangle|$ the effect for this small amount of $\mathcal{MP}$ is not visible. For a task sequencing problem, for example with eight manipulation primitives, the advantage of finding the optimal solution in comparison to use the nearest-neighbor solution do not really exist.

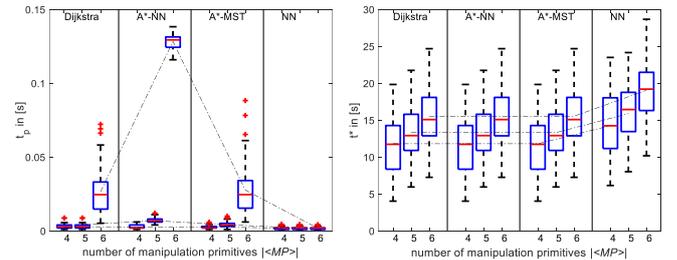

Fig. 6. Results task sequencing using different heuristics: results planning time (left); results task executing time (right).

Figure 5 summarizes the process of task and path planning in a coherent architecture. Based on the relational model created from CAD data, the processor computes the graph with the symbolic degrees of freedom, which is further used to generate a set of manipulation primitives. The information is then fused with the vision data and due to the concept of visibility and accessibility it is possible to get a local set of executable manipulation primitives. The local set is further sequenced. For executing the manipulation primitive set also path information is required. So, in the next section we will discuss a new and more efficient approach for global path planning.

*D. Path planning*

To compute a collision-free path $\underline{p} = \underline{f}(x, y, z, \varphi_x, \varphi_y, \varphi_z)$ for an entire fixed-base spatial manipulator with n-revolute joints, the configuration space $\mathcal{C}$ is a subset of $(\mathbb{R}^3 \times SO(3))^n$ [44], makes it practically impossible to compute it completely. To deal with this problem, many search- or sampling-based path planning algorithms are known, which implicitly compute $\mathcal{C}$. But still a major problem is the scaling of the computational complexity of path planning algorithms with the structure of the environment map. In case of an optimal algorithm $\mathcal{A}_{opt}$, the results scale directly with the exploration stepsize s. Consequently, it can be observed that there are two boundary cases, which means for a small exploration stepsize $s_\downarrow$ the planning results in a principally shorter path $\underline{p}_\downarrow$ and a high planning time $t_{p,\uparrow}$. The opposite case is true for a large $s_\uparrow$, compare (9), (10).

$$\underline{p}_\downarrow, t_{p,\uparrow} \leftarrow \mathcal{A}_{opt}(s_\downarrow). \quad (9)$$

$$\underline{p}_\uparrow, t_{p,\downarrow} \leftarrow \mathcal{A}_{opt}(s_\uparrow). \quad (10)$$

The problem is to find a suitable $s_\Delta$, for which it is possible to find a nearly optimal path $\underline{p}_\downarrow \approx \underline{p}_{opt}$ in a short planning time $t_{p,\downarrow}$. Also, the problem must be avoided that for $s_\uparrow$ no path will be found. To deal with this problem, we will develop a method, which uses well-known path planning algorithms and adapt the exploration stepsize automatically to the local environment conditions using the knowledge of the map occupancy. Therefore, we introduce a pre-processing of the environment map to compute well-suited stepsizes for all subspaces.

*1) Adaptive space division for global path planning*

Further, we will discuss the theoretical approach to adapt the exploration stepsize of the global path planning algorithms automatically to a given map. For map representation we use a probabilistic voxel map $\mathcal{PVM} \in \mathbb{R}^3$. Because the map is represented by a set of voxel $\bigcup_i^N v_i = \mathcal{PVM}$, which are described by the states {occupied o, free f, unknown u} $\in v_i$ it is possible to introduce a measure of the global occupancy $\rho_{global}$, described by

$$\rho_{global} = \frac{\sum_j o_j}{\sum_i v_i}, \text{with } j \leq i; \; \rho_{global} \in [0 \dots 1]. \quad (11)$$

The correlation of the exploration stepsize in respect to the map occupancy has been chosen as

$$s_{\Delta,\uparrow} \leftarrow \rho_\downarrow. \quad (12)$$

$$s_{\Delta,\downarrow} \leftarrow \rho_\uparrow. \quad (13)$$

With the adaption of $s_\Delta$, the planner can explore the path in free space with a large $s_{\Delta,\uparrow}$ and in more cluttered space with a smaller exploration stepsize $s_{\Delta,\downarrow}$. To consider also the local occupancy, the method divides $\mathcal{PVM}$ iterative in subspaces $\mathcal{R}_i$. This means that $\mathcal{PVM}$ represents a tree $\mathcal{T}$ and the corresponding subspaces $\mathcal{R}_i$ leafs. In every division, eight new subspaces are created per parent node in dependency of the local occupancy of $\mathcal{R}_i$, so that the map can be handled as an octree. Every node is described by a cuboid $C = I_x \times I_y \times I_z = [x_{min}, x_{max}] \times [y_{min}, y_{max}] \times [z_{min}, z_{max}]$ which is divided into eight equal leaf cubes $c_1, c_2 \dots, c_8$. The method is now straightforward, because the exploration stepsize is adapted independently of the depth of the tree $d_\mathcal{T}$, so

$$s_\Delta = g(d_\mathcal{T}) = g(f(\rho)). \quad (14)$$

Algorithm 2 and Figure 7 describes the complete adaptive space subdivision (ASD) procedure.

---

**Algorithm 2:** Adaptive Space Division (ASD)

**Input:** Probabilistic Voxel Map $\mathcal{PVM}$.

**Output:** Probabilistic Voxel Map with space subdivision $\mathcal{PVM}_{ASD}$ and set of stepsizes $\{s_\Delta\}$ for corresponding subspace $\mathcal{R}_i$

1: // *Compute global occupancy*
2: $\rho \leftarrow \frac{\sum_j o_j \in \mathcal{PVM}}{\sum_i v_i \in \mathcal{PVM}}$;
3: $\langle s_\Delta \rangle \leftarrow g(d_\mathcal{T})$;
4: $k = 1$;
5: // *Start Space Subdivision*
6: **for** $i = 1:1:8$
7:    **while** ($\rho > \rho_{limit}$) **do**
8:       **if** ($k > 1$) **then**
9:          $\rho \leftarrow \frac{\sum_j o_j \in \mathcal{R}_i}{\sum_i v_i \in \mathcal{R}_i}$;
10:         $s \leftarrow g(d_\mathcal{T})$;
11:       **end if**
12:       $(\mathcal{R}_{i,1}, \mathcal{R}_{i,2}, \dots, \mathcal{R}_{i,8}) \leftarrow \text{divideSpace}(\mathcal{R}_i)$;
13:       $\mathcal{PVM}_{ASD} \leftarrow \{\mathcal{R}_{i,1}, \mathcal{R}_{i,2}, \dots, \mathcal{R}_{i,8}\}$;
14:       $\langle s_\Delta \rangle \leftarrow s$;
15:       $k \leftarrow k + 1$;
16:    **end while**
17: **end for**
18: **return** ($\mathcal{PVM}_{ASD}, s_\Delta$);

1: **function** $(c_1, c_2 \dots, c_8) \leftarrow \text{divideSpace}(C)$
2:    $c_1 \leftarrow [x_{min} + 0.5|I_x|, x_{max}] \times [y_{min}, y_{min} + 0.5|I_y|] \times [z_{min}, z_{min} + 0.5|I_z|]$
3:    $c_2 \leftarrow [x_{min}, x_{min} + 0.5|I_x|] \times [y_{min}, y_{min} + 0.5|I_y|] \times [z_{min}, z_{min} + 0.5|I_z|]$
4:    $\vdots$
5:    $c_8 \leftarrow [x_{min}, x_{min} + 0.5|I_x|] \times [y_{min} + 0.5|I_y|, y_{max}] \times [z_{min} + 0.5|I_z|, z_{max}]$
6: **end function**

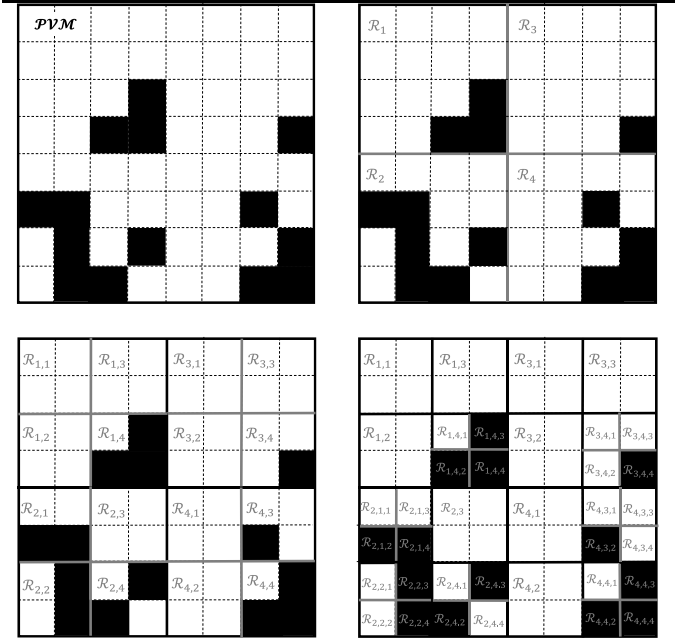

Fig. 7. Schematic two-dimensional representation of the adaptive space division algorithm. The building of the tree is done using iterative depth-first.

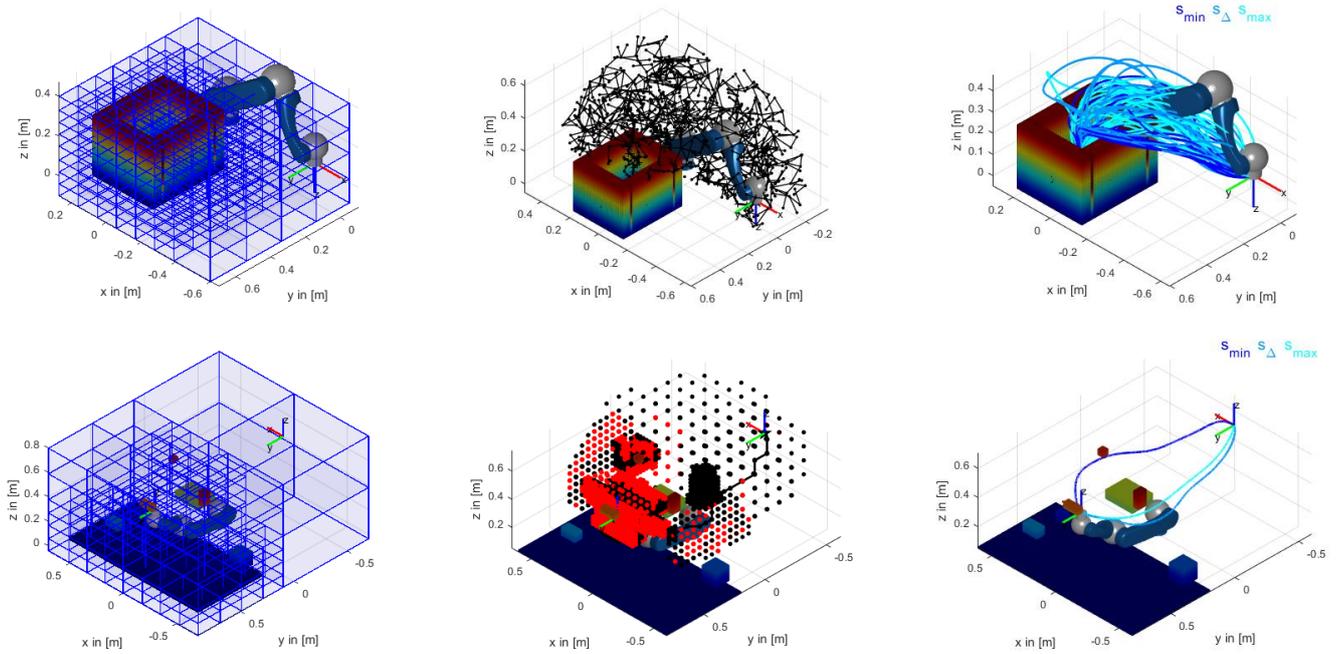

Fig. 8. Scenarios and results of the ASD approach: Results of the ASD algorithm scenario 1 (left/ first row) and scenario 2 (left/ second row); Exploration with RRT using scenario 1 (middle/ first row); Exploration with A* using scenario 2 (middle/ second row); Resulting path with adaptive, minimum and maximum exploration stepsize for scenario 1 with RRT (right/ first row) and scenario 2 with A* (right/ second row).

*2) Path planning algorithms*

For path planning we apply search-based planners with Greedy and A* and also sampling-based planners with a RRT and a bidirectional-RRT. The collision detection is done using a two-stage concept. The first stage verifies only a sphere to voxel detection and the second-stage uses oriented bounding boxes for every link. The kinematic is described through DH-parameters. To smooth the result of the path planner Bernstein-Bezier splines are used.

*3) Experimental results: path planning*

To evaluate the results of the adaptive subspace division, we compare it against the constant minimal exploration stepsize $s_{min} \in s_\Delta$ and the constant maximum exploration stepsize $s_{max} \in s_\Delta$. Two scenarios are used to validate the method. Scenario 1 is a typical industrial task like bin picking and scenario 2 consists of an occupied space through which the robot must move, see Figure 8 first column. Figure 8 also summarizes the results using an adaptive exploration stepsize in comparison to a constant one. The different exploration stepsizes according to the subspace occupancy can be seen clearly. It has been found experimentally that a subdivision of $d_\mathcal{T} = 3$ is sufficient and a further subdivision does not offer further computational advantages for the applied workspace.

The exploration stepsize is here computed linearly $s_{\Delta,d_\mathcal{T}} = 0.5 \cdot s_{\Delta,d_\mathcal{T}-1}$ with respect to the actual tree depth $d_\mathcal{T} \in \mathbb{N}_+$ and $s_{\Delta,0} = s_{max}$. To evaluate the approach typical metrics [45] are used with:

- path planning time $t_p$ in [s]
- collision detection time $t_c$ in [s]
- path length $|\underline{p}|$ in [m]
- deviation optimal path $|\Delta \underline{p}^*| \in \{x \in \mathbb{R}_+ | x \geq 1\}$
- path smoothness $\kappa$ in [rad]
- number of explored vertices $|V| \in \mathbb{N}_+$
- success rate $S \in [0 \dots 1]$ during one minute

The experiments are repeated five times for the search-based planners and thirty times for the sampling-based planners. Table II represents the results for scenario 1 and Table III for scenario 2, as the arithmetic mean over all experimental runs. Figure 9 shows the path planning time, the deviation from the optimal path and the explored nodes as box-plot.

In almost all cases of search-based planners, the path planning and the collision detection time can be reduced by up to 70% percent. Also in all cases a path can be computed in comparison to scenario 2 where it is not possible to find a path using $s_{max}$ and the Greedy algorithm.

The deviation from the optimal path is also in most cases better when the ASD algorithm is applied than the maximal exploration stepsize. In some cases, the path of the adaptive exploration stepsize $s_\Delta$ results in a larger path length than using $s_{max}$, because of the static discretization grid, which is constant in a subspace $\mathcal{R}_i$. Figure 10 shows this effect of the discretization according to the resulting path length.

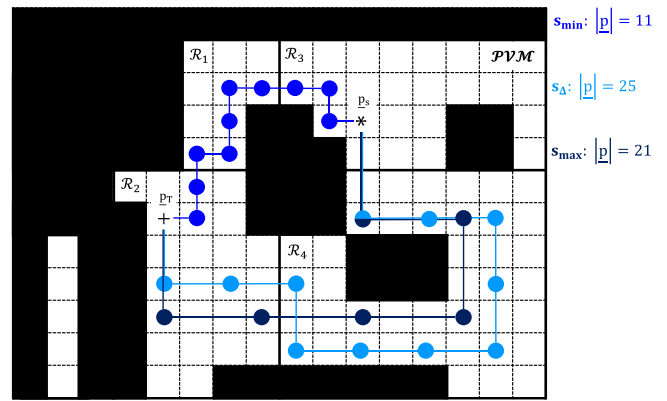

Fig. 10. Discretization with different exploration stepsizes: Using $s_{min}$ results in the minimal path length. With the adaptive exploration stepsize $s_\Delta$ the path length is greater than with the maximal exploration stepsize $s_{max}$ because the discretization grid of $s_{max}$ is closer to the boundaries of the obstacle.

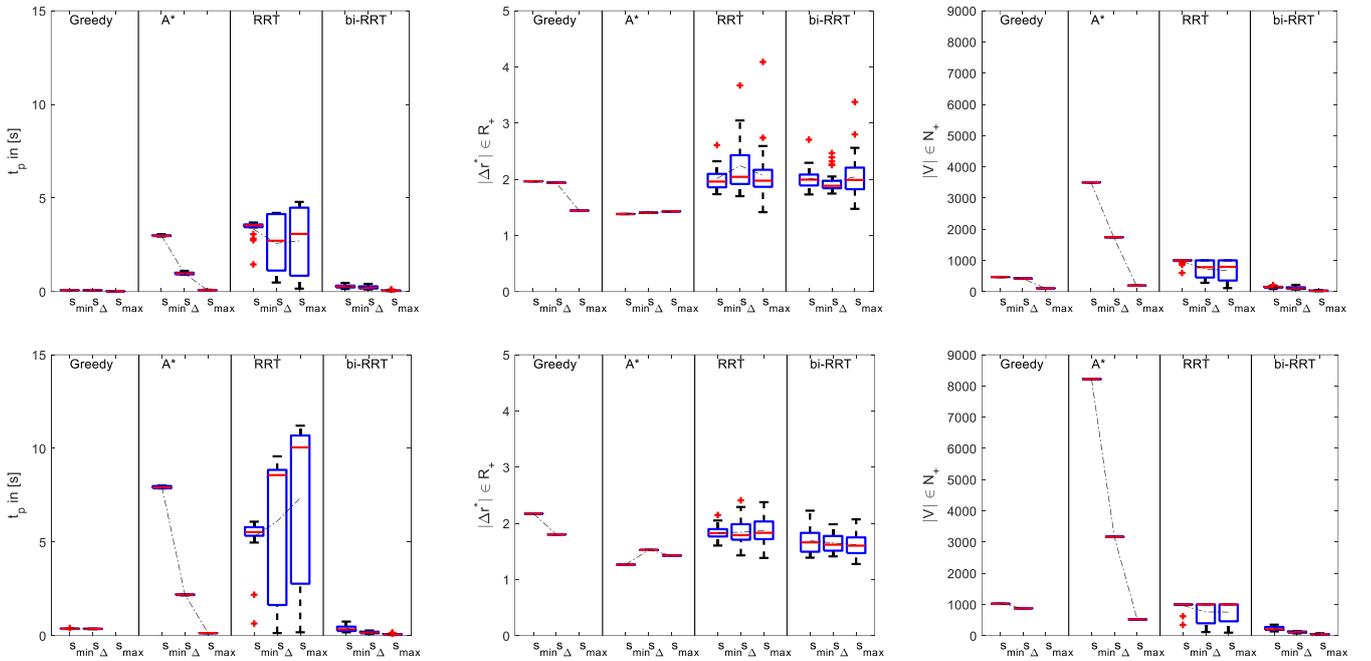

Fig. 9. Box-Plots of the metric for path planning time: scenario 1 (left/ first row), scenario 2 (right/ first row); deviation from optimal path: scenario 1 (middle/ second row), scenario 2 (middle/ second row); number of explored nodes: scenario 1 (right/ first row), scenario 2 (right/ second row).

TABLE II. EVALUATED METRICS FOR PATH PLANNING- SCENARIO 1

| Scenario 1- Box | Greedy ($s_{min}$) | Greedy ($s_\Delta$) | Greedy ($s_{max}$) | A* ($s_{min}$) | A* ($s_\Delta$) | A* ($s_{max}$) | RRT ($s_{min}$) | RRT ($s_\Delta$) | RRT ($s_{max}$) | bi-RRT ($s_{min}$) | bi-RRT ($s_\Delta$) | bi-RRT ($s_{max}$) |
|---|---|---|---|---|---|---|---|---|---|---|---|---|
| $t_p$ in [s] | 0.081 | 0.076 | 0.018 | 2.991 | 0.979 | 0.085 | 3.343 | 2.583 | 2.713 | 0.279 | 0.225 | 0.057 |
| $t_c$ in [s] | 1.344 | 1.647 | 0.326 | 34.48 | 14.04 | 1.229 | 23.35 | 17.39 | 17.34 | 8.491 | 6.804 | 1.685 |
| $\lvert \underline{p} \rvert$ in [m] | 1.325 | 1.308 | 0.973 | 0.932 | 0.947 | 0.963 | 1.358 | 1.512 | 1.396 | 1.358 | 1.316 | 1.378 |
| $\lvert \Delta \underline{p}^* \rvert \in \mathbb{R}_{+\backslash 1}$ | 1.965 | 1.939 | 1.443 | 1.381 | 1.404 | 1.428 | 2.013 | 2.242 | 2.070 | 2.014 | 1.951 | 2.043 |
| $\kappa$ in [rad] | 1.048 | 1.182 | 0.452 | 0.420 | 0.373 | 0.559 | 0.846 | 0.904 | 1.166 | 0.893 | 0.905 | 1.371 |
| $\lvert V \rvert \in \mathbb{N}_+$ | 471 | 422 | 109 | 3500 | 1744 | 193 | 965 | 727 | 675 | 144 | 113 | 29 |
| $S \in [0 \dots 1]$ | 1 | 1 | 1 | 1 | 1 | 1 | 0.63 | 0.73 | 1 | 1 | 1 | 1 |

TABLE III. EVALUATED METRICS FOR PATH PLANNING- SCENARIO 2

| Scenario 2- occ. Space | Greedy ($s_{min}$) | Greedy ($s_\Delta$) | Greedy ($s_{max}$) | A* ($s_{min}$) | A* ($s_\Delta$) | A* ($s_{max}$) | RRT ($s_{min}$) | RRT ($s_\Delta$) | RRT ($s_{max}$) | bi-RRT ($s_{min}$) | bi-RRT ($s_\Delta$) | bi-RRT ($s_{max}$) |
|---|---|---|---|---|---|---|---|---|---|---|---|---|
| $t_p$ in [s] | 0.374 | 0.365 | - | 7.929 | 2.188 | 0.141 | 5.291 | 6.111 | 7.354 | 0.384 | 0.170 | 0.075 |
| $t_c$ in [s] | 4.067 | 4.089 | - | 27.86 | 18.69 | 0.695 | 8.552 | 8.026 | 6.556 | 4.209 | 2.302 | 0.868 |
| $\lvert \underline{p} \rvert$ in [m] | 2.606 | 2.162 | - | 1.518 | 1.831 | 1.712 | 2.199 | 2.210 | 2.246 | 2.029 | 1.980 | 1.945 |
| $\lvert \Delta \underline{p}^* \rvert \in \mathbb{R}_{+\backslash 1}$ | 2.172 | 1.802 | - | 1.265 | 1.526 | 1.427 | 1.833 | 1.842 | 1.872 | 1.691 | 1.650 | 1.621 |
| $\kappa$ in [rad] | 1.466 | 2.023 | - | 0.372 | 0.396 | 0.613 | 0.807 | 0.893 | 1.138 | 0.749 | 0.810 | 1.059 |
| $\lvert V \rvert \in \mathbb{N}_+$ | 1022 | 878 | - | 8221 | 3168 | 520 | 968 | 760 | 756 | 225 | 116 | 47 |
| $S \in [0 \dots 1]$ | 1 | 1 | 0 | 1 | 1 | 1 | 1 | 1 | 1 | 1 | 1 | 1 |

The adaptive space subdivision is a good compromise for sampling-based planners, too. The number of explored vertices can be reduced in all cases and also the improvement of the success rate is possible through an adapted exploration stepsize.

## IV. CONCEPT FOR INTERFACING TASKS AND PATH PLANNING

In the previous sections the task and path planning have been described in detail. Obviously, the task sequence is the input information for the path planning algorithm. However, there is a discrepancy in how the information is presented for the two modules. The task sequencing operates on data in a symbolic form, whereas the path planner requires specific Cartesian data.

In order to assure a high degree of flexibility of the implementation, also for future research, the serialization of the manipulation primitives (output of task sequencing) has been carried out using a custom domain-specific language (DSL). This is defined as close to the theoretical description of manipulation primitives as possible. The interpreter for this DSL has been automatically generated based on the Extended Backus-Naur Form (EBNF) of the language syntax using the ANTLR parser generator. ANTLR can generate a parser based on the syntax definitions in a wide range of languages (e.g. C#, Java, C++), which can be integrated in the target system (e.g. as a ROS node). The task sequencer generates an application program written in this custom-defined DSL. Using the ANTLR generated parser, this application program can be interpreted on the target system. As a next goal for this ongoing research, the custom parser will be integrated in the robot controller. Based on the application program, the parser will call the skill primitives [31] of the robot with the correct parameterization. Thus, the parser will link the symbolic manipulation primitives with skill primitives, which are a proven paradigm for the execution of complex robot tasks. Further advantages of using a DSL are:

- Close representation of the mathematical formalism of general task planning
- Flexible serialization of commands and even logic if needed
- Human-readable application program (similar for all control equipment in a factory setting)

## V. CONCLUSION AND FUTURE WORK

In this work an approach for task and path planning was investigated, which allows a robot system to autonomously plan different maintenance tasks. Using a novel sampling-based approach to compute the relative degree of freedom between components in combination with the symbolic concept of manipulation primitives, a powerful method for task planning is available. Further, a universal concept for reducing the planning time of global path planners is provided. An actual limitation of the task planning approach is that it works only for a single-arm manipulator. In future it will be extended so that it works also for dual-arm manipulation. Also we will focus on the development of the parser, which enables interfacing planning and execution.